\documentclass[conference]{IEEEtran}
\IEEEoverridecommandlockouts
\usepackage{cite}
\usepackage{amsmath,amssymb,amsfonts}
\usepackage{algorithmic}
\usepackage{graphicx}
\usepackage{textcomp}
\usepackage{xcolor}

\usepackage{multirow}
\usepackage{enumitem}

\def\BibTeX{{\rm B\kern-.05em{\sc i\kern-.025em b}\kern-.08em
    T\kern-.1667em\lower.7ex\hbox{E}\kern-.125emX}}
    
\begin{document}

\title{Data Standardization for Robust Lip Sync}

\author{\IEEEauthorblockN{Chun Wang}
\IEEEauthorblockA{ \\
\textit{Mashang Consumer Finance Co., Ltd.}\\
 Chongqing, China \\
lukewang25@live.cn}
}

\maketitle

\begin{abstract}
Lip sync is a fundamental audio-visual task.
However, existing lip sync methods fall short of being robust in the wild.
One important cause could be distracting factors on the visual input side, making extracting lip motion information difficult.
To address these issues, this paper proposes a data standardization pipeline to standardize the visual input for lip sync.
Based on recent advances in 3D face reconstruction, we first create a model that can consistently disentangle lip motion information from the raw images.
Then, standardized images are synthesized with disentangled lip motion information, with all other attributes related to distracting factors set to predefined values independent of the input, to reduce their effects.
Using synthesized images, existing lip sync methods improve their data efficiency and robustness, and they achieve competitive performance for the active speaker detection task.
\end{abstract}
\section{Introduction}
\label{sec:intro}
Lip sync is a fundamental audio-visual (AV) task.
Its primary function is to determine when the audio and visual streams are out of sync.
For this reason, it is frequently formulated as a cross-modal matching task and is resolved by comparing the representations of two modalities \cite{perfectMatch}.
Active speaker detection (ASD), whose goal is to detect which or no subject in visual streams is the speaker, can be solved using such a formulation.
A solid baseline for resolving ASD can be created by performing lip sync on each subject who presents in the video \cite{ASW2021}.

Lip sync performance, however, degrades in real-world conditions.
One major cause is the very diverse nature of videos taken in the wild, with the majority of the diversity caused by distracting factors that could make extracting lip motion information difficult.
For example, when faces are in large head poses, performance significantly decreases \cite{chung2017lip}\cite{cheng2020towards}.

There are primarily two types of strategies for improving the robustness of AV methods.
One is a data-driven strategy.
Previous research, for example, has shown that training with a large number of non-frontal faces progressively makes AV systems more robust to big head poses \cite{chung2017lip}\cite{cheng2020towards}.
However, adopting an entirely data-driven strategy would make lip sync data-hungry and hard to optimize, as a significant amount of data covering a wide range of combinations would be needed to address compound distracting factors.

\begin{figure}[tb]
  \vspace{-1.0em}
  \begin{minipage}[b]{1.0\linewidth}
    \centering
    \centerline{\includegraphics[width=8.5cm]{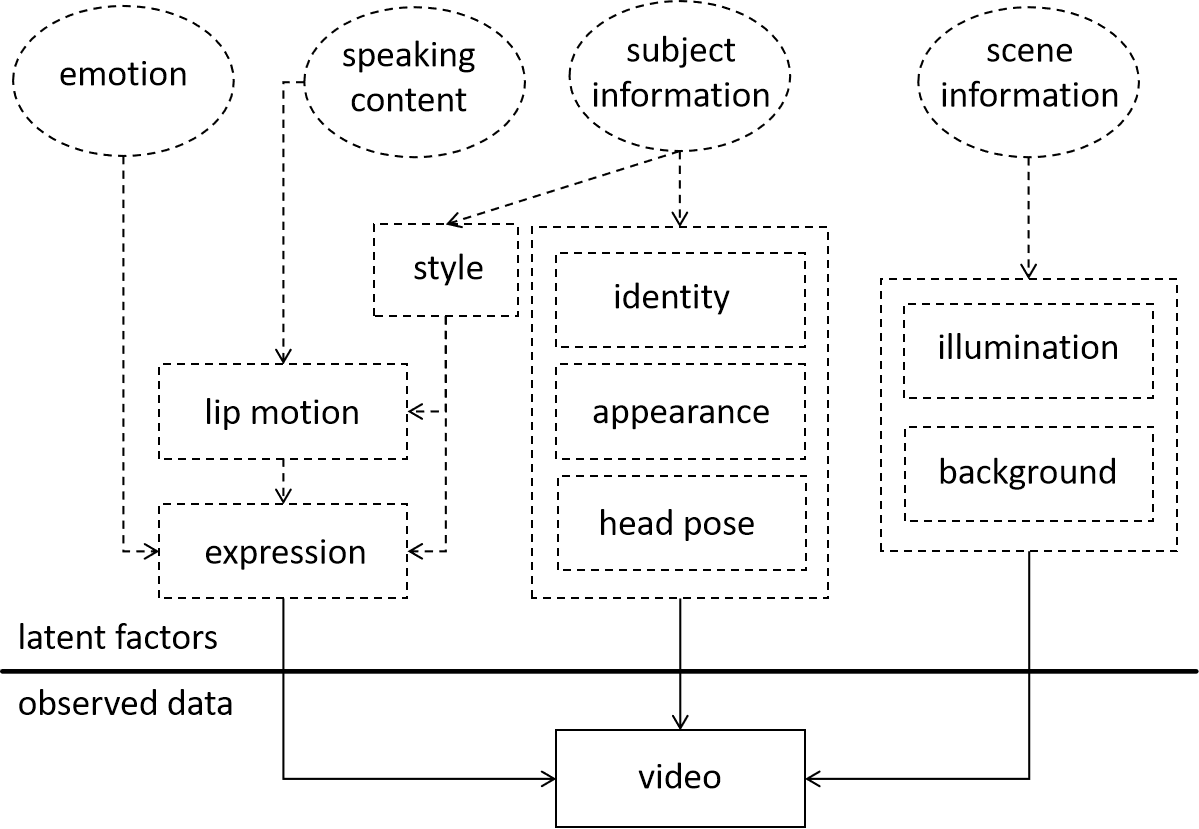}}
  \end{minipage}
  \caption{A domain knowledge-based model of the main factors that affect videos, with miscellaneous factors like occlusion omitted.}
  \label{fig:causal}
  \vspace{-1.0em}
\end{figure}

As a complement, others take a more domain-knowledge-based strategy.
For instance, several face frontalization techniques \cite{kang2021robust}\cite{koumparoulis2018deep} are proposed as one way to explicitly reduce the effects of head poses.
According to domain knowledge, see Fig.~\ref{fig:causal}, in addition to head poses, subject-related factors (primarily identity and appearance) and scene-related factors (primarily illumination and background) should also be regarded as distracting factors, for the reason that judgments of AV synchronization should not be influenced by their variations.
Meanwhile, lip motion information is the primary visual cue for lip sync. 
Methods that can disentangle lip motions from compound distracting factors are thus desired.


Lip motions can be viewed as a form of facial expression in a broader sense.
ExpNet \cite{ExpNet} and DeepExp3D \cite{DeepExp3D} demonstrate that the 3D Morphable Model (3DMM) \cite{egger20203d} may be used to acquire disentangled expressions.
The 3DMM defines a parametric space where expression is unrelated to other attributes.
In particular, this expression subspace is demonstrated to capture speaking-related facial motions well \cite{DeepExp3D}.
Moreover, the image formation model (see Sec.~\ref{sec:preliminaries}) built on top of 3DMM provides a larger parametric space with additional control over attributes that are related to many factors.

This paper proposes a data standardization pipeline (DSP) that can produce standardized expressive images with the effects of compound distracting factors reduced by leveraging the image formation model.
First, a network, named E-Net, is developed to consistently disentangle expression from the input at the video level. 
Then, using the image formation model, expressions disentangled from the input are used to synthesize expressive images, with all other attributes corresponding to distracting factors set to predefined values, independent of the input, to reduce their effects on the synthesized images.
Experiments demonstrate that by taking images standardized by the DSP as input, existing lip sync methods increase their data efficiency and generalizability and achieve competitive performance for the ASD task on the recent ASW dataset \cite{ASW2021}.
The rest of the paper is organized as follows:
Sec.~\ref{sec:relatedwork} discusses related works.
Then, building on the preliminary information provided in Sec.~\ref{sec:preliminaries}, the DSP is then thoroughly described in Sec.~\ref{sec:Approach}.
The experimental settings and results are presented in Sec.~\ref{sec:Experiments}.
Finally, Sec.~\ref{sec:conclusion} concludes the paper with a discussion on future improvements.

\section{RELATED WORK}
\label{sec:relatedwork}
\textbf{3DMM coefficients}.
Several ambiguities make estimating 3DMM coefficients difficult \cite{egger20203d}.
Major expression-related ambiguity results from the fact that certain changes in facial shape can be ascribed to either identity variations or expressions, which can be inferred from \eqref{eq:S}.
In a recent benchmark \cite{REALY}, Deep3D \cite{Deng_2019_CVPR_Workshops} and MGCNet \cite{MGCNET} are shown to fit the mouth area well.
But they exhibit inconsistent ascriptions across images as they attempt to estimate all attributes using preimage features.
To resolve the ambiguity, some \cite{tewari2019fml}\cite{Mallikarjun2021learn} propose learning an improved version of 3DMM with more orthogonal identity and expression subspaces.
Instead of enhancing the 3DMM model, some researchers prefer to introduce constraints during optimization. 
For instance, by aggregating multiple images of a subject and assuming that they are of the same underlying subject-specific attributes, \cite{Thies_2016_CVPR} optimizes global attributes specific to the subject, and attributes image-varying facial morphing to the expression.
Despite being simpler, optimization-based methods are less robust in the wild \cite{ExpNet}.
In this paper, we integrate this global constraint into Deep3D \cite{Deng_2019_CVPR_Workshops} to improve disentanglement consistency.
Note that in this work, we are focused on improving the consistency of expression disentanglement at the video level, rather than on more accurate 3D face reconstruction.

\textbf{Face synthesis/generation}. 
There are various methods for face frontalization.
Some use generative models \cite{koumparoulis2018deep} to translate a profile face to a frontal one.
\cite{Zhou_2020_CVPR} blends generative models with 3DMM for 3D face rotations.
However, they may not preserve the expression of the input image \cite{kang2021robust} and generative models are less robust when generating videos.
Others, instead, propose to synthesize expression-preserving frontal faces via warping \cite{kang2021robust}.
However, those methods are specifically designed to address head poses, and adapting them to manage compound factors could be challenging.
Otherwise, it has been shown that full parametric models (see Sec.~\ref{sec:preliminaries}) can be used to synthesize faces with diversified attribute combinations \cite{Deng_2019_CVPR_Workshops}\cite{MGCNET}.
This method involves both controllability and steady synthesis quality.
In this paper, we explore its potential in handling compound distracting factors.

\textbf{Lip sync and ASD}. 
For lip sync, SyncNet \cite{chung2017lip} is a reliable baseline.
PerfectMatch \cite{perfectMatch} enhances performance by introducing a multi-way matching approach.
With regard to ASD, diverse viewpoints exist.
The most widely used AVA dataset \cite{AVA2020} just demands the association of audio and visual data, whereas the more recent ASW dataset \cite{ASW2021} requires the synchronization of two modalities, which excludes instances like dub-subbed movies.
As it would be impossible to handle dubbings using the lip motion cue alone, advanced relational modeling \cite{Kopuklu_2021_ICCV} is required.
Hence, we resolve to the ASW.

\section{PRELIMINARIES}
\label{sec:preliminaries}
\textbf{The 3DMM}. 
In 3DMM, the textured 3D face model $\boldsymbol{F}(\boldsymbol{S}, \boldsymbol{T})$ is defined by the face shape $\boldsymbol{S}$ and the face texture $\boldsymbol{T}$, and both are represented as parametric linear models:
\begin{align}
	   \label{eq:S}
        \boldsymbol{S}(\boldsymbol{\alpha}, \boldsymbol{\beta}) &= \boldsymbol{\bar{S}} + \boldsymbol{B}_{id}\boldsymbol{\alpha} + \boldsymbol{B}_{exp}\boldsymbol{\beta} \\
        \label{eq:T}
	   \boldsymbol{T}(\boldsymbol{\delta}) &= \boldsymbol{\bar{T}} + \boldsymbol{B}_{tex}\boldsymbol{\delta}
\end{align}
where $\boldsymbol{\bar{S}}$ and $\boldsymbol{\bar{T}}$ are the mean face shape and texture; $\boldsymbol{B}_{id}$, $\boldsymbol{B}_{exp}$, and $\boldsymbol{B}_{tex}$ are PCA bases for identity, expression, and texture, respectively. 
Specifically, Basel Face Model 2009 \cite{BFM09} and FaceWarehouse \cite{FaceWarehouse} built bases are adopted, thus identity $\boldsymbol{\alpha} \in \mathbb{R}^{80}$, expression $\boldsymbol{\beta} \in \mathbb{R}^{64}$, and texture $\boldsymbol{\delta} \in \mathbb{R}^{80}$.

\textbf{The illumination model}. 
Under the distant smooth illumination assumption and the Lambertian skin surface assumption, the illumination is modeled via Spherical Harmonics(SH) \cite{SH}, and per-vertex radiosity is
\begin{equation}
\boldsymbol{t}^\prime_{i}(\boldsymbol{n}_{i}, \boldsymbol{t}_{i} | \boldsymbol{\gamma}) = \boldsymbol{t}_{i}\sum_{b=1}^{B^{2}} \gamma_{b} \Phi_{b} (\boldsymbol{n}_{i})
\end{equation}
where $\boldsymbol{n}_{i}$ is the normal, $\boldsymbol{t}_{i}$ is the texture color, $\Phi_{b}$ are SH basis functions. 
We choose $B$=3 for each of the red, green, and blue channels, resulting in illumination coefficients $\boldsymbol{\gamma} \in \mathbb{R}^{27}$.

\textbf{The image formation model}. The pose $\boldsymbol{p} \in \mathbb{R}^{6}$ is represented by a rotation $\boldsymbol{R} \in SO(3)$, parameterized with three Euler angles, and a translation $\boldsymbol{t} \in \mathbb{R}^{3}$. 
Thus, the image formation process can be presented as 
\begin{equation}
\boldsymbol{I}(\boldsymbol{\alpha}, \boldsymbol{\beta},\boldsymbol{\delta},\boldsymbol{\gamma}, \boldsymbol{p}) = \Pi(\boldsymbol{R}\boldsymbol{F}(\boldsymbol{S}, \boldsymbol{T}^\prime) + \boldsymbol{t})
\label{eq:render}
\end{equation}
where $\Pi: \mathbb{R}^{3} \rightarrow \mathbb{R}^{2} $ is the perspective projection matrix; and $\boldsymbol{T}^\prime$ is the illuminated 3D face texture.

By setting proper coefficients, we can set faces and environments accordingly and synthesize images of corresponding attributes through graphic rendering, e.g. rasterization.
See \cite{egger20203d} for a thorough introduction.

\section{APPROACH}
\label{sec:Approach}
In this section, we introduce the design requirements and implementations of the proposed DSP. 

\begin{figure}[tb]
\vspace{-1.0em}
\begin{minipage}[b]{1.0\linewidth}
  \centering
  \centerline{\includegraphics[width=8.5cm]{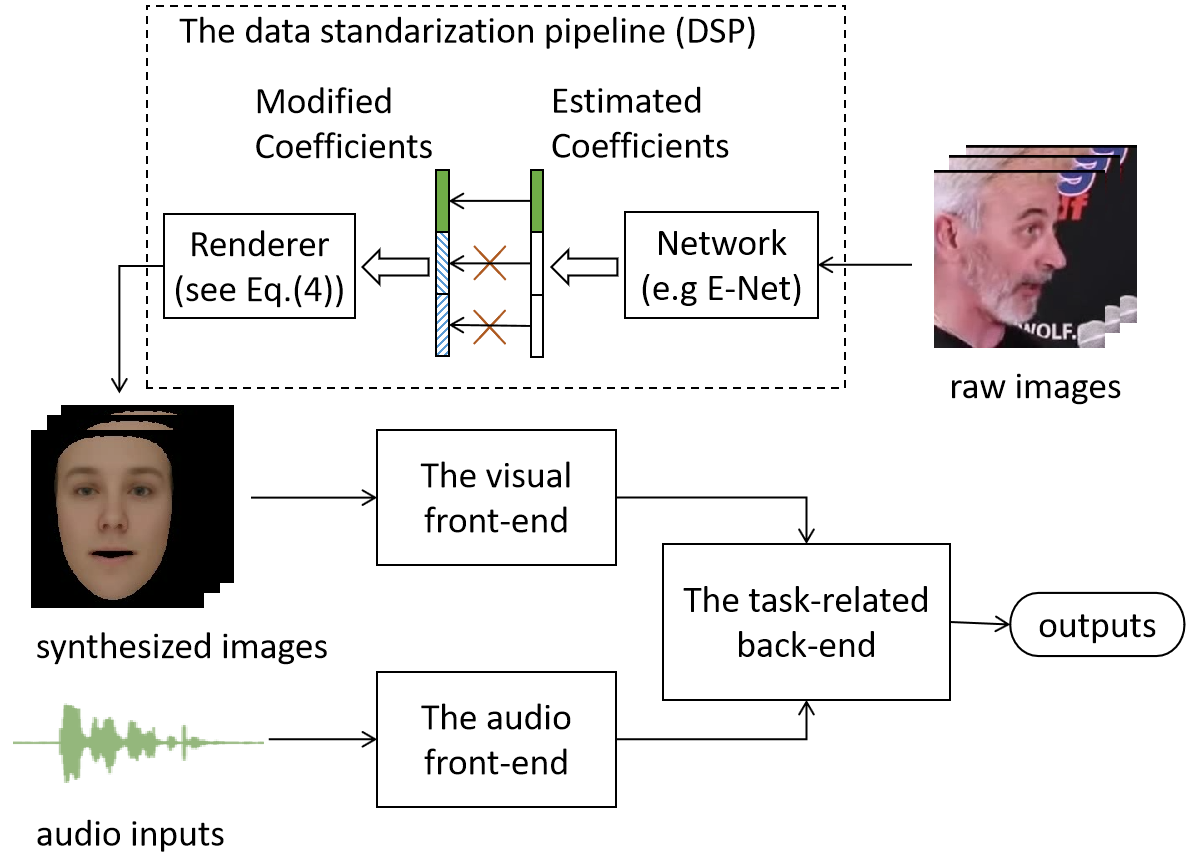}}
\end{minipage}
\caption{Lip sync with DSP schematic.}
\label{fig:schematic}
\vspace{-1.0em}
\end{figure}

\subsection{Overview of DSP}
\label{ssec:Overview}
As depicted in Fig.~\ref{fig:schematic}, the visual front-end, audio front-end, and task-related back-end are the three parts that make up modern lip sync model architectures \cite{perfectMatch}.
The DSP should be placed between the raw images and the visual front-end. 
It takes raw images as inputs and produces standardized images, which are then fed into the visual front-end. 
Further, the DSP is built using the parametric image formation model described in Sec.~\ref{sec:preliminaries} and thus has two components: a network for estimating the required coefficients from the input, and a renderer that uses the coefficients to synthesize images.
Their implementations should fulfill the design requirements.
\begin{enumerate}
\setlength{\itemsep}{0pt}
\setlength{\parsep}{0pt}
\setlength{\parskip}{0pt}
\item The DSP is robust on its own in the wild.
\item Preserving lip motion information from the raw images.
\item Reducing compound distracting factors.
\item No significant new distracting factors are introduced.
\end{enumerate}
The image formation model is based on physical rules, which are stable in the wild, hence the DSP meets requirement 1.
To satisfy requirement 2, a network that focuses on consistently resolving expression-related ambiguities is customized by modifying \cite{Deng_2019_CVPR_Workshops}, see Sec.~\ref{ssec:TrainingDatasetConstruction} to Sec.~\ref{ssec:TrainingStrategy}.
To satisfy requirement 3, standardized images are sythesized with modified coefficients, see Sec.~\ref{ssec:FaceSynthesis}.
Regarding requirement 4, missing teeth in synthesized images might indeed introduce a new distracting factor. 
However, experimental results presented in Sec.~\ref{sec:Experiments} suggest that this has negligible negative effects on lip sync, as the cues for lip motion remain intact.

\subsection{Training Dataset Construction}
\label{ssec:TrainingDatasetConstruction}
For training, several image collections are necessary.
First, a large number of single-speaker videos are gathered, and then each video is processed for face detection/tracking and 3D landmark detection \cite{Deng_2019_CVPR_Workshops} in order to create face tracks \cite{ASW2021}.
Each face track is a sequential sequence of a subject's facial images, and a subject may show up in more than one face track.
Then, around $M$ facial images are sampled from each of the high-quality face tracks \cite{DeepExp3D}.
Finally, the dataset contains $K$ collections, and each has $M_{k}$ pairs of facial images and landmarks $\{l_{n}\}$ of a subject, i.e. $\boldsymbol{C}_{k}=\left\{ \left( \boldsymbol{I}_{i},  \{l_{n}\}_{i}\right) \right\}_{i=1}^{M_{k}}$.

We use speaking videos for the following reasons:
1) Facial images from short-duration videos are more likely to fulfill the assumption that they are of the same underlying subject-specific attributes.
2) They enhance training by introducing diversified lip motions and implicit temporal stability. 

\begin{figure}[tb]

\begin{minipage}[b]{1.0\linewidth}
  \centering
  \centerline{\includegraphics[width=8.5cm]{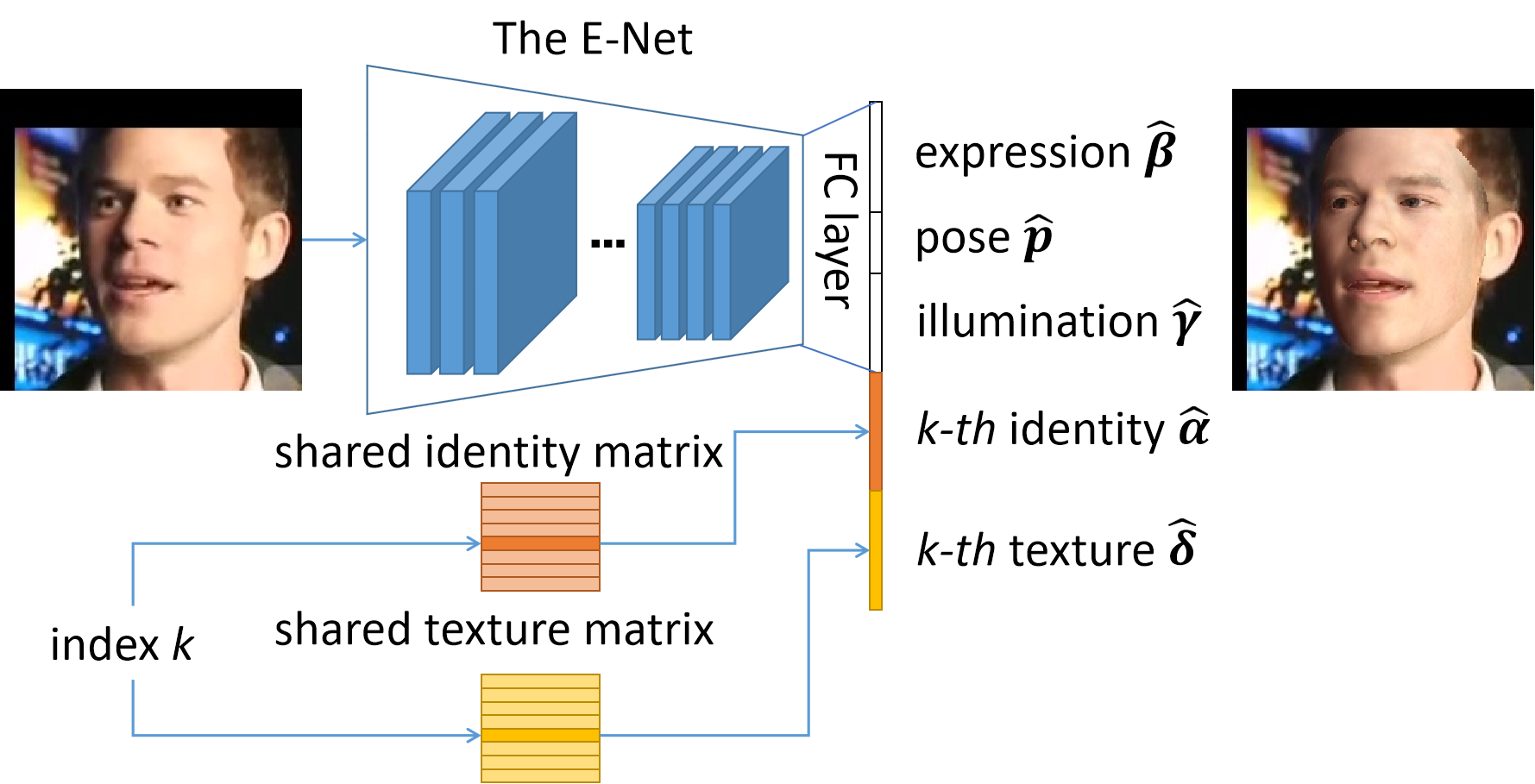}}
  \centerline{(a) Model architecture}\medskip
\end{minipage}
\hfill
\begin{minipage}[b]{1.0\linewidth}
  \centering
  \centerline{\includegraphics[width=8.5cm]{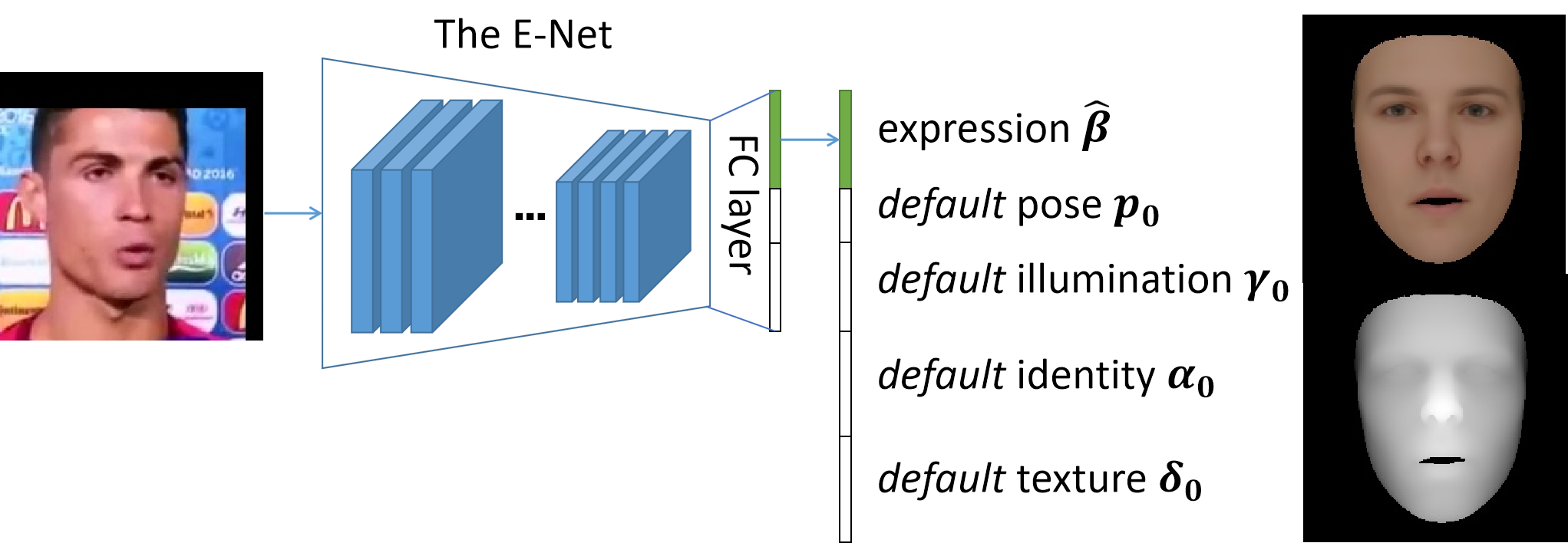}}
  \centerline{(b) Face synthesis}\medskip
\end{minipage}
\caption{Overview of the proposed approach. }
\label{fig:exp}
\vspace{-1.0em}
\end{figure}

\subsection{Model Architecture}
\label{ssec:ExpressionExtractionModelArchitecture}
The model consists of a regression network and two trainable matrices, as depicted in Fig.~\ref{fig:exp}(a).
The estimation of preimage attributes is carried out using the regression network, termed as E-Net.
Similar to \cite{Deng_2019_CVPR_Workshops} \cite{MGCNET}, the E-Net is implemented as a ResNet-50 \cite{He_2016_CVPR} with the last fully-connected layer of 97 dimensions, for regressing a 97-dimensional vector of preimage attributes ($\boldsymbol{\beta} \in \mathbb{R}^{64}$, $\boldsymbol{\gamma} \in \mathbb{R}^{27}$, $\boldsymbol{p} \in \mathbb{R}^{6}$).
One of those trainable matrices is the shared identity matrix of dimension 80-by-$K$, where 80 is the dimension of identity ($\boldsymbol{\alpha} \in \mathbb{R}^{80}$) and $K$ is the number of collections in the dataset. 
It is denoted as shared because the identity shared by all of the facial images in the $k$-th collection is represented in its $k$-th column.
Similarly, the other is the shared texture matrix of dimension 80-by-$K$ since $\boldsymbol{\delta} \in \mathbb{R}^{80}$. 

\subsection{Loss Functions}
\label{ssec:LossFunctions}

To measure the discrepancy for an estimated coefficient $\boldsymbol{\hat{x}}=\{ \boldsymbol{\hat{\alpha}}_{k}, \boldsymbol{\hat{\delta}}_{k}, \boldsymbol{\hat{\beta}}, \boldsymbol{\hat{p}}, \boldsymbol{\hat{\gamma}} \}$, different losses are used, following \cite{Deng_2019_CVPR_Workshops}.

\textbf{The photometric loss}. It measures the photometric consistency between the input $\boldsymbol{I}$ and the synthesized one $\boldsymbol{I}'$.
\begin{equation}
L_{pho}\left(\boldsymbol{x}\right) =\dfrac{\sum _{i\in \boldsymbol{\mathcal{M}}}\boldsymbol{A}_{i}\cdot \left\| \boldsymbol{I}_{i} - \boldsymbol{I}_{i}'\left( \boldsymbol{x}\right) \right\| _{2}}{\sum _{i \in \boldsymbol{\mathcal{M}} }\boldsymbol{A}_{i}}
\end{equation}
where $\boldsymbol{\mathcal{M}}$ and $\boldsymbol{A}$ are two masks indicating the facial area and the per-pixel skin probability \cite{Deng_2019_CVPR_Workshops}, respectively.

\textbf{The landmarks distance (LMD) loss}. It measures the geometric consistency between detected landmarks $\{l_{n}\}$ and ones $\{l_{n}'\}$ reprojected from predefined 3D shape vertices \cite{Deng_2019_CVPR_Workshops}. 
\begin{equation}
L_{lan}\left( \boldsymbol{x}\right) =\dfrac{1}{N}\sum ^{N}_{n=1}\omega _{n}\left\| l_{n}-l_{n}'\left(\boldsymbol{x}\right) \right\| _{2}
\label{eq:lrl}
\end{equation}
where $\omega _{n}$ is the weight of the $n$-th landmark. 
Lip landmarks are weighted to 10, otherwise 1.

\textbf{The regularization loss}. It adds zero-mean Gaussian distribution prior constraints on 3DMM coefficients. 
\begin{equation}
L_{reg}\left(\boldsymbol{x}\right) = \omega _{\boldsymbol{\alpha}} \left\|\boldsymbol{\alpha}\right\| ^{2} + \omega _{\boldsymbol{\beta}} \left\|\boldsymbol{\beta}\right\| ^{2} + \omega _{\boldsymbol{\delta}} \left\|\boldsymbol{\delta}\right\| ^{2} 
\end{equation}
where $\omega _{\boldsymbol{\alpha}}=0.5$, $\omega _{\boldsymbol{\beta}}=2.0$ and $\omega _{\boldsymbol{\delta}}=0.5$. 

In total, all terms are weighted combined as 
\begin{equation}
    \begin{aligned}
        L\left(\boldsymbol{x}\right) = &\lambda_{pho}L_{pho}\left(\boldsymbol{x}\right) + \lambda_{lan}L_{lan}\left( \boldsymbol{x}\right) + \lambda_{reg}L_{reg}\left(\boldsymbol{x}\right)
    \end{aligned}
\end{equation}
where $\lambda_{pho}=1.92$, $\lambda_{lan}=1.6e{-3}$, and $\lambda_{reg}=3.0e{-4}$.

Most hyper-parameters are directly adopted from \cite{Deng_2019_CVPR_Workshops}, with stronger regularization on the expression term to prevent the identification of all variations in facial shapes as expressions, thereby making the identity attribute trivial \cite{tewari2019fml}.
Note that we do not explicitly regularize the disentanglement between texture and illumination, as they will be replaced during face synthesis. 
However, a similar idea can be applied.

\subsection{Training Strategy}
\label{ssec:TrainingStrategy}
The whole model, including the E-Net and two trainable matrices, is trained jointly with built collections in a way similar to \cite{Deng_2019_CVPR_Workshops}, with the following modifications made. 
1) As depicted in Fig.~\ref{fig:exp}(a), instead of being regressed, subject-specific attributes are trainable parameters indexed from shared matrices and shall be updated during training. 
2) For a batch, we sample multiple collections first, then several data from each collection.
Data from the same collection constrain one another and assist in resolving ambiguities as they share the same subject-specific attributes but differing preimage ones.

\subsection{Face Synthesis}
\label{ssec:FaceSynthesis}
We propose to synthesize a standardized face with depth as inputs for subsequent AV tasks.
As described in Sec.~\ref{ssec:Overview}, it is expected that the effects of compound distracting factors will be reduced in the synthesized images while the expression information from the raw input will be preserved.
This can be done by setting appropriate coefficients to the image formation model, as stated in Sec.~\ref{sec:preliminaries}.

In specific, we utilize modified coefficients for rendering instead of all coefficients estimated from the raw input.
As depicted in Fig.~\ref{fig:exp}(b), the trained E-Net estimates the expression singly from the input. 
Meanwhile, attributes corresponding to distracting factors are explicitly set to predefined default values, regardless of inputs.
By rendering with modified coefficients, the synthesized images are devoid of the effects of compound distracting factors while preserving the expression from the raw images.
For the ease of subsequent AV tasks, we set uniform white illumination, mean identity, mean texture, and frontal pose as defaults, so that $\boldsymbol{\hat{x}}=\{ \boldsymbol{\alpha}=\boldsymbol{0}, \boldsymbol{\delta}=\boldsymbol{0}, \boldsymbol{\hat{\beta}},\boldsymbol{\gamma}=\boldsymbol{\gamma}_{\boldsymbol{0}}, \boldsymbol{p}=\boldsymbol{p}_{\boldsymbol{0}} \}$, and render an expressive image $\boldsymbol{I}'(\boldsymbol{\hat{x}})$.
Examples are given in Fig.~\ref{fig:illustration}, where raw images (column a) vary greatly whereas synthesized images (column e) are standardized and free of distracting factors (pose, illumination, etc.).

We can also render a corresponding pseudo depth map $\boldsymbol{D}'$ in order to more accurately describe 3D lip motions.
This is done by rendering with $\boldsymbol{Z}'_{\boldsymbol{S}}$ instead of illuminated $\boldsymbol{T}'$ as the texture (see \eqref{eq:render}), where $\boldsymbol{Z}'_{\boldsymbol{S}}$ is the normalized Z-coordinator of $\boldsymbol{S}$,
\begin{equation}
\boldsymbol{Z}'_{\boldsymbol{S}} = \frac{\boldsymbol{Z}_{\boldsymbol{S}} - \boldsymbol{Z}_{min}}{\boldsymbol{Z}_{max} - \boldsymbol{Z}_{min}},
\end{equation}
where $\boldsymbol{Z}_{min}=\min(\boldsymbol{Z}_{\boldsymbol{S}_{i}})$ and $\boldsymbol{Z}_{max}=\max(\boldsymbol{Z}_{\boldsymbol{S}_{i}})$ are the min depth and max depth of the video clip, respectively.

Overall, we synthesize standardized expressive RGB/RGBD images from raw facial images.
Notably, we also reduce the effect of the often poor image quality found in raw data via synthesis.

\section{EXPERIMENTS}
\label{sec:Experiments}
\subsection{Datasets}
\label{ssec:dataset}
We develop the proposed model and the lip-sync model using VoxCeleb2 \cite{chung2018voxceleb2}.
Its dev-split has over 1 million utterances from 145569 videos of 5994 subjects and its test-split has 36237 utterances from 4911 videos of 118 subjects.

To develop the proposed model, we collect 5000 subjects from the dev-split.
Ten utterances are sampled from each of the two videos for each subject, resulting in $K=50k$ collections with each having about $M=50$ samples.
Separately, a minidev-split consisting of 5000 utterances from 500 subjects is sampled from the dev-split for training the lip-sync model, while the entire test split is used for evaluation.
Note that the minidev-split is less than ${1}/{7}$ the size of the test-split.

Regarding the ASD task, we directly evaluate lip-sync models models trained with VoxCeleb2 on the test split of ASW dataset \cite{ASW2021}.
The test split consists of 53 videos (51 accessible), with 4.5 hours of active (i.e., sync) face tracks and 3.4 hours of inactive ones.

\begin{figure}[tb]

\begin{minipage}[b]{1.0\linewidth}
  \centering
  \centerline{\includegraphics[width=8.5cm]{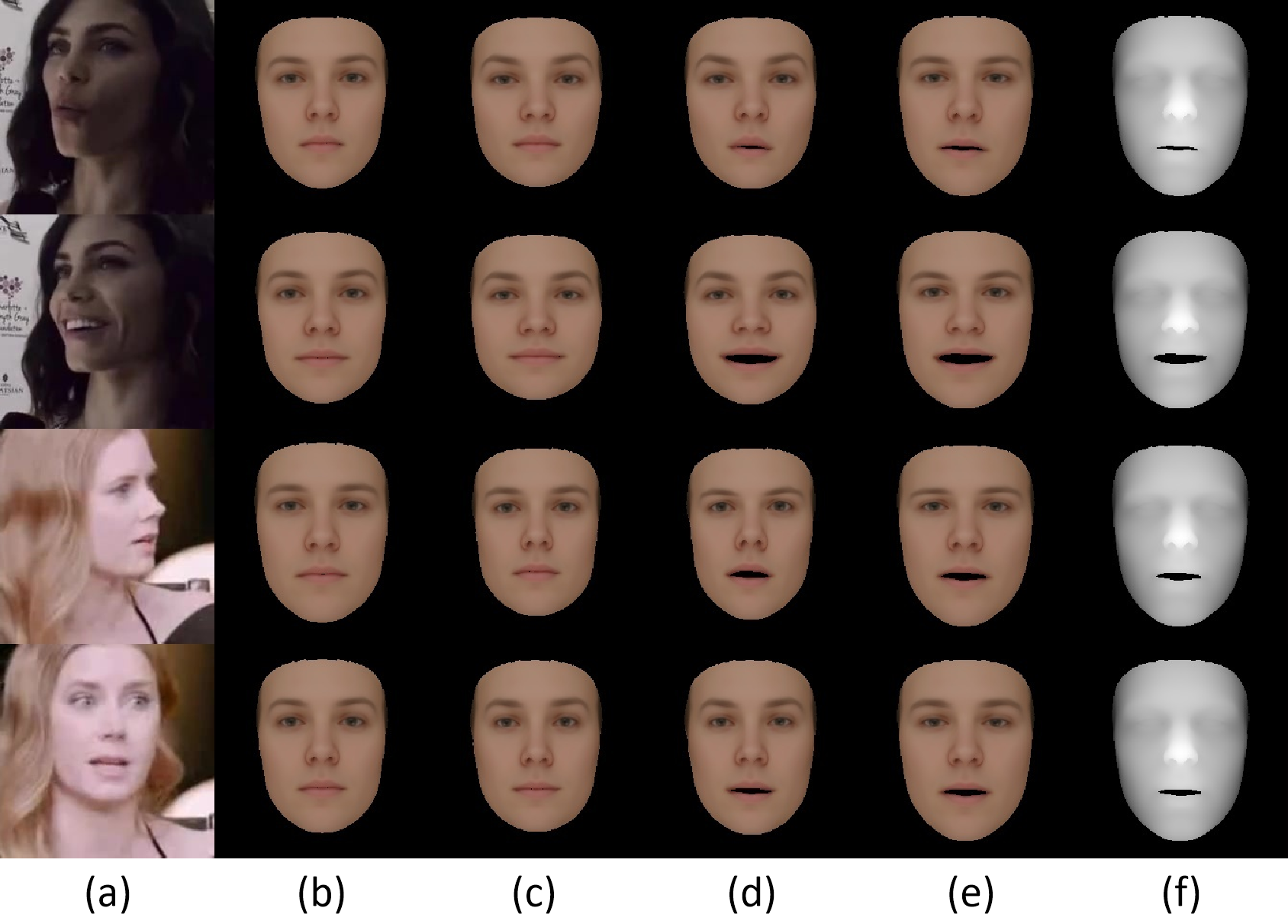}}
\end{minipage}
\caption{Snapshots of synthesized data. (a) raw images; (b)-(d) synthesized RGB, with MGCNet estimated id, Deep3D estimated id, and Deep3D estimated id and exp, respectively; (e)-(f) synthesized RGBD with E-Net estimated exp.}
\label{fig:illustration}
\vspace{-1.0em}
\end{figure}

\begin{table}[tb]
\vspace{-1.0em}
    \centering
    \caption{Lip-fit accuracies. Lower is better.}
    \begin{tabular}{|l|l|lll|}
    \hline
    \multirow{2}{*}{\textbf{Method}} &
      \multirow{2}{*}{\textbf{\begin{tabular}[c]{@{}l@{}}Attributes\end{tabular}}} &
      \multicolumn{3}{c|}{\textbf{Group LMD}} \\ \cline{3-5} 
                            &             & \multicolumn{1}{l|}{\textbf{min}} & \multicolumn{1}{l|}{\textbf{max}} & \textbf{avg.} \\ \hline
    GT                      & $\boldsymbol{\alpha}_{gt}$, $\boldsymbol{\beta}_{gt}$   & \multicolumn{1}{l|}{1.085} & \multicolumn{1}{l|}{1.093}  & 1.091  \\ \hline
    \multirow{2}{*}{Deep3D \cite{Deng_2019_CVPR_Workshops}} 
                            & $\boldsymbol{\alpha}_{gt}$, $\boldsymbol{\hat{\beta}}$  & \multicolumn{1}{l|}{2.948} & \multicolumn{1}{l|}{2.975}  & 2.967  \\
                            & $\boldsymbol{\hat{\alpha}}$, $\boldsymbol{\hat{\beta}}$ & \multicolumn{1}{l|}{2.339} & \multicolumn{1}{l|}{2.369}  & 2.358  \\ \cline{2-5} \hline
    \multirow{2}{*}{MGCNet \cite{MGCNET}} 
                            & $\boldsymbol{\alpha}_{gt}$, $\boldsymbol{\hat{\beta}}$  & \multicolumn{1}{l|}{3.353} & \multicolumn{1}{l|}{3.372}  & 3.365   \\
                            & $\boldsymbol{\hat{\alpha}}$, $\boldsymbol{\hat{\beta}}$ & \multicolumn{1}{l|}{2.902} & \multicolumn{1}{l|}{2.920}  & 2.912   \\ \cline{2-5} \hline
    \multirow{1}{*}{Ours E-Net}              
                            & $\boldsymbol{\alpha}_{gt}$, $\boldsymbol{\hat{\beta}}$  & \multicolumn{1}{l|}{\textbf{1.235}} & \multicolumn{1}{l|}{\textbf{1.241}} & \textbf{1.240} \\ \hline
    \end{tabular}
\label{tab:lmd}
\vspace{-1.0em}
\end{table}

\subsection{Results on Expression Disentanglement}
\label{ssec:LipmotionsDisentanglement}
According to Sec.~\ref{ssec:TrainingStrategy}, we first train the proposed model.
More specifically, two matrices are initialized with Gaussian noise $\mathcal{N}(\mu=0,\,\sigma=0.01)$ and the E-Net is ImageNet-pretrained \cite{russakovsky2015imagenet}.
Two collections and eight samples from each are sampled for each batch.
With a constant learning rate of $1.0e{-4}$, the model is trained using the Adam optimizer for 20 epochs.

Due to the lack of datasets with ground truth (GT) expressions of $\boldsymbol{\beta} \in \mathbb{R}^{64}$, we use 5-fold cross-validation for the evaluation.
Particularly, $50k$ collections are split into 5 groups.
Each time, we train with four groups and evaluate the holdout group, whose GT coefficients are obtained by overfitting it with the proposed method.
We use the LMD (see \eqref{eq:lrl}) on lip landmarks as a metric.
The group LMD is defined as the mean of the collection LMDs, and each is defined as the mean of the data LMDs.
We form mixed coefficients in order to isolate the effects of certain estimated attributes.
For instance, the mixed coefficients are formed as $\boldsymbol{\hat{x}}=\{ \boldsymbol{\alpha}_{gt}, \boldsymbol{\delta}_{gt}, \boldsymbol{\hat{\beta}}, \boldsymbol{p}_{gt}, \boldsymbol{\gamma}_{gt} \}$ in order to evaluate the estimated expression.

We compare with Deep3D \cite{Deng_2019_CVPR_Workshops} and MGCNet \cite{MGCNET}.
The min, max, and average group LMDs are reported in Tab.~\ref{tab:lmd}, and results with GTs are also included as references.
For Deep3D and MGCNet, results with estimated expressions perform worse than counterparts with GT expressions, and additionally performing image-wise identity estimation can improve results.
These results suggest that they incorrectly ascribe some parts of face morphings to identity variations (see Fig.~\ref{fig:illustration}(b)(c)), and as a result, their estimated expressions when used alone cannot adequately describe lip motions.
Our results using the E-Net estimated expressions, otherwise, are more in line with those using GTs, showing that the E-Net can more consistently attribute facial morphings to expressions.

\begin{table}[tb]
\vspace{-1.0em}
\centering
\caption{Lip-sync accuracies of different methods.}
\begin{tabular}{|l|l|l|l|l|}
\hline
\textbf{Input Mode}     &\textbf{Attributes}  & \textbf{Train} & \textbf{Eval} & \textbf{Acc.}    \\ \hline
Raw data              &none                     & dev            & test          & 94.1\%          \\ \hline
Raw data              &none                     & minidev        & test          & 88.9\%          \\ \hline
RGB(\cite{Deng_2019_CVPR_Workshops}) & $\boldsymbol{\hat{\alpha}}$, $\boldsymbol{\hat{\beta}}$   & minidev        & test          & 97.4\%          \\ \hline
RGBD(\cite{Deng_2019_CVPR_Workshops}) & $\boldsymbol{\hat{\alpha}}$, $\boldsymbol{\hat{\beta}}$   & minidev        & test          & 98.1\%          \\ \hline
RGB(ours)           & $\boldsymbol{\hat{\beta}}$ 				  & minidev        & test          & 99.1\% \\ \hline
RGBD(ours)          & $\boldsymbol{\hat{\beta}}$ 				  & minidev        & test          & \textbf{99.2\%} \\ \hline
\end{tabular}
\label{tab:lipsyncacc}
\vspace{-1.0em}
\end{table}

\begin{table}[tb]
\vspace{-1.0em}
\centering
\caption{ASD performances of different methods.}
\begin{tabular}{|l|l|l|l|}
\hline
\textbf{Method}                         	  & \textbf{AP} & \textbf{AUROC} & \textbf{EER} \\ \hline
Self-supervised \cite{ASW2021}\cite{ASW2021arXiv}           	  & 0.924       & 0.962          & 0.083        \\ \hline
RGBD(\cite{Deng_2019_CVPR_Workshops} )        & 0.929       & 0.954         & 0.109         \\ \hline
RGBD(ours)              	 		       &  \textbf{0.957}       & \textbf{0.971}         & \textbf{0.079}         \\ \hline
\end{tabular}
\label{tab:asdperformance}
\vspace{-1.0em}
\end{table}

\subsection{Results on Audio-Visual Tasks}
\label{ssec:results}
We conduct experiments on lip sync and ASD tasks.
We adopt the self-supervised PerfectMatch \cite{perfectMatch} for lip sync, using synthesized images as the visual input and Mel-frequence cepstral coefficients (MFCCs) as the audio input.
We first train a model on the dev-split of VoxCeleb2 using the raw face tracks as a baseline for comparisons.
Then, a variety of versions are trained using the minidev split, including raw face tracks, synthesized images with E-Net estimated expressions, and synthesized images with estimated identities and expressions from Deep3D.

For evaluation, we follow \cite{perfectMatch} to extract audio and visual features from every 0.2s segment with a stride of 0.04s.
Then, we calculate the cosine similarity between each visual feature and each audio feature within a $\pm 15$ frame window, and we determine the offset providing the min distance.
If the determined offset is within $\pm 1$ frame of the GT offset, it is correct.

The evaluation results are listed in Tab.~\ref{tab:lipsyncacc}.
With less data, models trained with synthesized images outperform those trained with raw face tracks.
Further, models trained with RGBD images yield better results.
This may be the case because depth information makes 3D lip motions more distinct, see Fig.~\ref{fig:illustration}.
Particularly, models trained with E-Net estimated expressions consistently outperform others, suggesting that improved lip motion description may directly benefit lip sync.


In the ASD task, effective lip-sync models are directly applied to ASW's test split \cite{ASW2021} without training or finetuning on ASW. 
Following \cite{ASW2021}, an AV pair exceeding a threshold distance is deemed active. 
Our pipeline preprocesses face tracks, synthesizes images, and averages cosine similarities across 15 frames for robustness as suggested in \cite{perfectMatch}. 
Results in Tab.~\ref{tab:asdperformance} reveal that lip-sync models exhibit great generalizability by matching the strong baseline \cite{ASW2021} in all metrics, including average precision (AP), the area under the receiver operating characteristic (AUROC), and equal error rate (EER). 
This evaluation is challenging as our models are evaluated across datasets and trained with less data.

\section{CONCLUSION}
\label{sec:conclusion}

This paper introduces a DSP for lip sync, which is customized using 3D face reconstruction techniques. 
This customization provides flexibility for incorporating domain knowledge and enables the handling of compound distracting factors. 
Preliminary results suggest that by synthesizing standardized expressive images with disentangled lip motion information, the DSP enhances the data efficiency and robustness of existing lip sync methods. 

In the future, this conceptual DSP could be enhanced with more advanced 3D face reconstruction techniques. 
To be specific, an improved 3D face model such as FLAME \cite{Li_2017_ACM} might be more expressive and potentially capture expressions more accurately. 
Additionally, instead of using rendered RGB/RGBD images, directly utilizing expressive 3D face mesh as visual representations could lead to a better implementation of DSP, as more information is retained. 
Finally, we would like to emphasize that in the data-driven era, domain knowledge remains valuable. 
Correctly utilizing domain knowledge may aid other AV tasks that face similar challenges, such as lip reading \cite{Peng_2023_ACM}, emotion recognition \cite{Pei_2021_TMM}, and digital human \cite{Danvevecek_2023_ACM}, particularly when the collection and sanity-checking of multi-modality datasets is difficult.

\bibliographystyle{IEEEtran}
\bibliography{IEEEabrv,icme2023template}

\end{document}